\pdfoutput=1

\documentclass[11pt]{article}

\usepackage{emnlp2021}

\usepackage{times}
\usepackage{latexsym}

\usepackage[T1]{fontenc}

\usepackage[utf8]{inputenc}

\usepackage{microtype}

\usepackage{xspace}
\usepackage{siunitx,booktabs}
\usepackage{multirow}
\usepackage{graphicx}
\usepackage{mathtools}
\usepackage{amssymb}
\usepackage{tcolorbox}
\usepackage{pagecolor}
\usepackage{amsmath}
\usepackage[toc,page]{appendix}
\usepackage{enumitem}
\setlist{nosep}
\usepackage{makecell}
\usepackage{xcolor}
\usepackage{pifont}
\usepackage{subfigure}
\usepackage{url}
\usepackage{hyperref}

\title{ Understanding Politics via Contextualized Discourse Processing }

\author{ Rajkumar Pujari \and Dan Goldwasser\\
  Department of Computer Science\\
  Purdue University\\
  \texttt{\{rpujari,dgoldwas\}@purdue.edu}}

\begin{document}

\maketitle

\begin{abstract}
Politicians often have underlying agendas when reacting to events. Arguments in contexts of various events reflect a fairly consistent set of agendas for a given entity. In spite of recent advances in Pretrained Language Models, those text representations are not designed to capture such nuanced patterns. In this paper, we propose a Compositional Reader model consisting of encoder and composer modules, that captures and leverages such information to generate more effective representations for entities, issues, and events. These representations are contextualized by tweets, press releases, issues, news articles, and participating entities. Our model processes several documents at once and generates composed representations for multiple entities over several issues or events. Via qualitative and quantitative empirical analysis, we show that these representations are meaningful and effective.
\end{abstract}

\section{Introduction}\label{sec:intr}
Over the last decade political discourse has moved from traditional outlets to social media. This process, starting in the '08 U.S. presidential elections, has peaked in recent years, with former-president Trump announcing the firing of top officials as well as policy decisions over Twitter. This presents a new challenge to the NLP community, \textit{how can this massive amount of political content be used to create principled representations of politicians, their stances on issues and legislative preferences?}
 
This is not an easy challenge as in political texts perspective is often subtle rather than explicit \citep{plain-sight-2019}. Choices of mentioning or omitting certain entities or attributes can reveal the author's agenda. For example, tweeting ``\textit{mass shootings are due to a huge mental health problem}'' in reaction to a mass shooting is likely to be indicative of opposing gun control measures, despite the lack of an explicit stance in the text.

Recent advances in Pretrained Language Models (PLMs) in NLP \citep{bert-2019, xlnet-2019, roberta-2019} have greatly improved word representations via contextualized embeddings and powerful transformer units, however such representations alone are not enough to capture nuanced biases in political discourse. Two of the key reasons are: (i) they do not directly focus on entity/issue-centric data and (ii) they only represent linguistic context rather \textit{external} political context.

Our main insight is that effectively detecting such bias from text requires modeling the broader political context of the document. This can include understanding relevant facts related to the event addressed in the text, the ideological leanings and perspectives expressed by the author in the past, and the sentiment/attitude of the author towards the entities referenced in the text. We suggest that this holistic view can be obtained by combining information from multiple sources, which can be of varying types, such as news articles, social media posts, quotes from press releases and historical beliefs expressed by politicians.

\begin{figure}[!htbp]
    \centering
    \includegraphics[width=5.6cm]{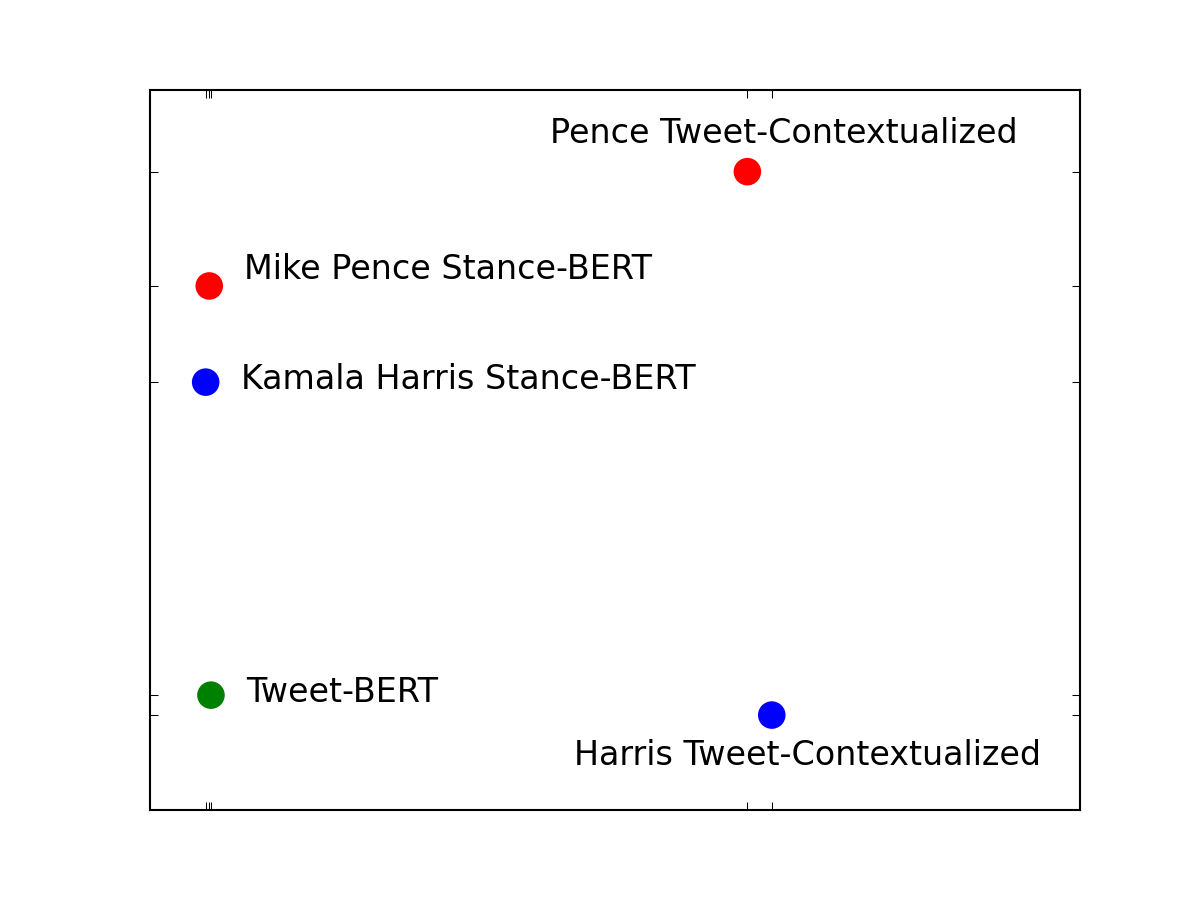}
    \caption{BERT vs. Author-Contextualized Encoder Composer Representation of an Ambiguous Tweet}
    \label{fig:intro_example}
\end{figure}

For example, consider the following tweet in context of a school shooting: \textit{We need to treat our teachers better! We should keep them safe.} If the author of the tweet is Kamala Harris (known to be pro-gun control), this tweet is likely to be understood as ``\textit{ban guns to avoid mass shootings in schools}''. However, if the same tweet is from Mike Pence, whose stance on \textit{guns} is: ``\textit{firearms in the hands of law abiding citizens makes our communities safer}'', the tweet could mean ``\textit{arming school teachers stops active shooters}''. This example demonstrates that depending on the context, the same text could signal completely different real-world actions. Hence, we need to model the broader context of the text in order to understand its true meaning. Visualization projecting the tweet representation into a 2D space is given in figure \ref{fig:intro_example}, and shows how contextualization from our model helps disambiguate this example. First, we show the BERT-base representation of the tweet ({\small\texttt{Tweet-BERT}}). We also show the BERT-base representations of the known stances of Pence and Harris on gun control ({\small\texttt{\{Mike Pence,Kamala Harris\} Stance-BERT}}). Finally, we apply our model, contextualizing the ambiguous tweet representation with speaker information ({\small\texttt{\{Mike Pence,Kamala Harris\} Tweet-Contextualized}}). The visualization captures how this representation can disambiguate the different interpretations of the same text, and capture their differences.

 A computational setting for this approach, \textit{combining text and context analysis}, requires two necessary attributes: (i) an \textit{input} representation that combines all the different types of information meaningfully and (ii) the ability to create \textit{a meaningful unified representation} in one-shot, that captures the complementary strengths of the different inputs. 
 
  We address the first challenge by introducing a graph structure that ties together first-person informal (tweets) and formal discourse (press releases and perspectives), third-person current (news) and consolidated (Wikipedia) discourse. These documents are connected via their authors, the issues/events they discuss and the entities mentioned in them. As a clarifying example consider the tweet by former-President Trump \textit{``The NRA is under siege by Cuomo''}. This tweet will be represented in our graph by connecting the text node to the author node (Trump) and the referenced entity node (NY Gov. Cuomo). These settings are shown in Fig. \ref{fig:test_graph}.

 We propose a novel neural architecture that unifies all the information in the graph in one-shot. \textit{Our architecture generates a distributed representation for each item in the graph that is contextualized by the representations of others}. It can dynamically respond to queries, focusing the induced representation on a specific context. In our example, this results in a modified tweet representation helping us characterize Trump's opinion of Cuomo \textit{in the context of the guns issue}. Our architecture consists of an \textit{Encoder} combining all documents related to a given node to generate an initial node representation and a \textit{Composer}, a Graph Attention Network (GAT), composing the graph structure to generate contextualized node embeddings.
 
We design two self-supervised learning tasks to train the model and capture structural dependencies over the rich discourse representation, predicting \textit{Authorship} and \textit{Referenced Entity} links over the graph structure. Intuitively, the model is required to understand subtle language usage; \textit{Authorship} prediction requires the model to differentiate between: (i) the language of one author from another and (ii) the language of the author in context of one issue vs another issue. \textit{Referenced Entity} prediction requires understanding the language used by a specific author when discussing a particular entity, given the author's past discourse.

We focus on a specific graph element--\textit{politicians}, and evaluate their resulting discourse representation on several empirical tasks which capture their stances and preferences. Our evaluation demonstrates the importance of each component of our model and usefulness of the learning tasks. To summarise, our research contributions include:

\begin{enumerate}[leftmargin=*,noitemsep]
    \item A novel graphical structure connecting various types of documents, entities, issues and events.
    \item An effective neural architecture, \textit{Compositional Reader}, processing all information in one-shot, and designing two effective tasks for training it.
    \item Designing \& performing quantitative and qualitative evaluation showing that our graph structure, neural architecture and learned representations are meaningful and effective for representing politicians and their stances on issues.\footnote{\href{https://github.com/pujari-rajkumar/compositional_learner}{Repository: https://github.com/pujari-rajkumar/compositional\_learner}}
\end{enumerate}

\section{Related Work}\label{sec:relw}
Due to recent advances in text representations catalysed by \citet{elmo-2018}, \citet{transformer-2017} and followed by \citet{bert-2019}, \citet{roberta-2019} and \citet{xlnet-2019}, we are now able to create rich textual representations, effective for many NLP tasks. Although contextual information is captured by these models, they are not explicitly designed to capture entity/event-centric information. Hence, tasks that require such information \citep{bias_prediction-2016,moral_foundation_2012,johnson-goldwasser-2016-identifying,vote_prediction-2018,perspectrum-2019}, would benefit from more focused representations.

Of late, several works attempted to solve such tasks, such as analyzing relationships and their evolution \citep{rmn-2016,larn-2019}, analyzing political discourse on news and social media \citep{21_mass_shootings_2019,roy-goldwasser-2020-weakly} and political ideology \citep{language_ideology-2012,beyond-binary-2017,political_ideology-2018}. Various political tasks such as roll call vote prediction \citep{roll-call-2008,kornilova-etal-2018-party,roll-call-2019,legislative-embeddings-2020,davoodi-etal-2020-understanding}, entity stance detection \citep{ tweet-stance-2016,stance-prediction-2019}, hyper-partisan/fake news detection \citep{li-goldwasser-2019-encoding,fake-news-semeval-2019,baly-etal-2020-written} require a rich understanding of the context around the entities that are present in the text.  But, the representations used are usually limited in scope to specific tasks and not rich enough to capture information that is useful across several tasks.

\indent The Compositional Reader model, that builds upon \citet{bert-2019} embeddings and consists of a transformer-based Graph Attention Network inspired from \citet{gat-2017} and  \citet{transformer-gat-2019}, aims to address those limitations via a generic entity-issue-event-document graph, which is used to learn highly effective representations.

Representing legislative preferences is typically done by modeling the ideal point of legislators represented in a Euclidean space from roll-call records~\cite{poole1997congress}. Recent approaches incorporate  bill text information into this representation~\cite{gerrish2011predicting,nguyen2015tea,kraft2016embedding,kornilova2018party}. Most relevant to our work is~\cite{spell2020embedding} which uses social media information. We significantly extend these approaches by contextualizing the social media content using a novel architecture.

\section{Data}\label{sec:data}
\begin{table}[!htbp]
    \centering
    \resizebox{!}{35pt}{
        \begin{tabular}{lc|lc}
        \hline
        \textbf{Data} & \textbf{Count} & \textbf{Data} & \textbf{Count} \\\hline
            News Events & $367$ & Tweets & $86,409$ \\
            Author Entities & $455$ & Press Releases & $62,257$\\
            Ref. Entities & $10,506$ & Perspectives & $30,446$\\
            Wikipedia & $455$ & News Articles & $8,244$\\\hline
            \textbf{Total Docs} & \textbf{$187,811$}\\\hline
        \end{tabular}
    }
    \caption{Summary statistics of data}
     \label{tab:data_stats}
\end{table}
We collected US political text related to $8$ broad topics: \textit{guns, LGBTQ rights, abortion, immigration, economic policy, taxes, middle east \& environment}. The data focused on $455$ members of the US Congress. We collected political text data relevant to above topics from $5$ sources: press statements by political entities from ProPublica Congress API\footnote{\url{https://projects.propublica.org/api-docs/congress-api/}}, Wikipedia articles describing political entities, tweets by political entities (\href{https://github.com/alexlitel/congresstweets}{Congress Tweets}, \citet{trump-tweets-2019}), perspectives of the senators and congressmen regarding various political issues from \href{https://www.ontheissues.org/}{ontheissues.org} and news articles \& background of the those political issues from \href{https://www.allsides.com/}{allsides.com}. A total of $187,811$ documents were used to train our model, as shown in Tab. \ref{tab:data_stats}.
\subsection{Event Identification}\label{sebsec:event_iden}
\indent To identify news events, we use news article headlines. We find the mean ($\mu$) and standard deviation ($\sigma$) of the number of articles published per day for each issue. If more than $\mu+\sigma$ number of articles are published on a single day for a given issue, we identify it as the beginning of an event. Then, we skip $7$ days and look for a new event.

\indent In our setting, events \textit{within} an issue are non-overlapping. We divide events for each issue separately, hence events for different issues do overlap. These events last for $7-10$ days on average and hence the non-overlapping assumption within an issue is a reasonable relaxation of reality. To illustrate our point: coronavirus and civil-rights are separate issues and hence have overlapping events. An example event related to coronavirus could be ``First case of COVID-19 outside of China''. Similarly an event about civil-rights could be ``Officer part of George Floyd killing suspended''. We inspected the events manually by random sampling. More example events are in the appendix.

\subsection{Data Pre-processing}\label{subsec:pre_proc}
We use Stanford CoreNLP tool \citep{corenlp-2014}, Wikifier \citep{wikifier-2017} and BERT-base-uncased implementation by \citet{hugging-face-transformers-2019} to preprocess data for our experiments. We tokenize the documents, apply coreference resolution and extract referenced entities from each document. The referenced entities are then wikified using Wikifier tool \citep{wikifier-2017}. The documents are then categorized by issues and events. News articles from \href{https://www.allsides.com/}{allsides.com} and perspectives from \href{https://www.ontheissues.org/}{ontheissues.org} are already classified by issues. We use keyword based querying to extract issue-wise press releases from Propublica API. We use hashtag based classification for tweets. A set of gold hashtags for each issue was created and the tweets were classified accordingly\footnote{Data collection is detailed in appendix}. Sentence-wise BERT-base embeddings of all documents are computed.
\subsection{Query Mechanism}\label{subsec:query_mech}
We implemented a query mechanism to obtain relevant subsets of data from the corpus. Each query is a triplet of \textit{entities, issues \& lists of event indices corresponding to each of the issues}. Given a query triplet, news articles related to the events for each of the issues, Wikipedia articles for each of the entities, background descriptions of the issues, perspectives of each entity regarding each of the issues and tweets \& press releases by each of the entities related to the events in the query are retrieved. Referenced entities for each of the sentences in documents and sentence-wise BERT embeddings of the documents are also retrieved. 

\section{Compositional Reader}\label{sec:comp}
\begin{figure}[!tb]
    \centering
    \includegraphics[width=5cm]{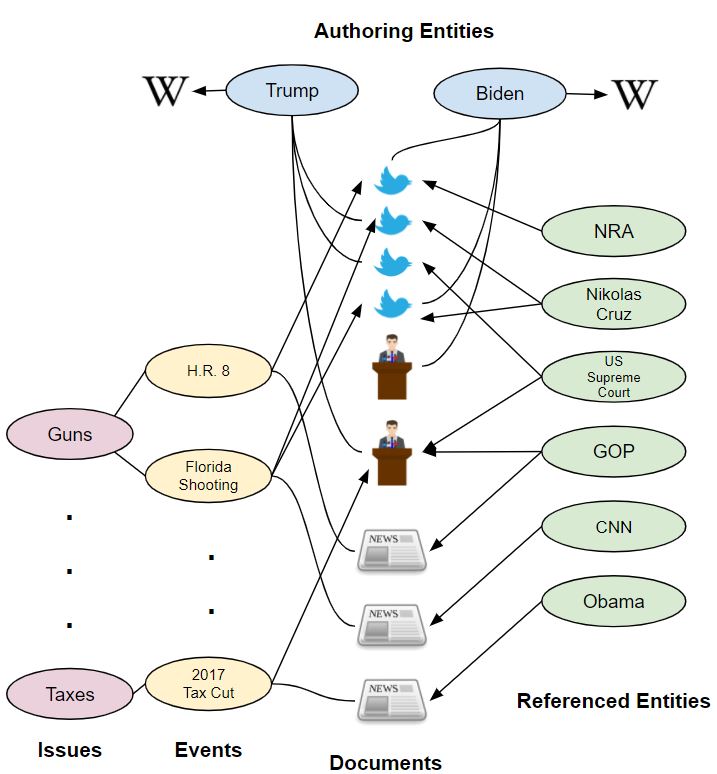}
    \caption{Example Text Graph from Graph Generator}
    \label{fig:test_graph}
\end{figure}
 In this section, we describe the architecture of the proposed `Compositional Reader' model in detail. It contains $3$ key components: Graph Generator, Encoder and Composer. Given a query output of the query mechanism from Sec. \ref{subsec:query_mech}, Graph Generator creates a directed graph with entities, issues, events and documents as nodes. Encoder is used to generate initial node embeddings for each of the nodes. Composer is a transformer-based Graph Attention Network (GAT) followed by a pooling layer. It generates the final node embeddings and a single summary embedding for the query graph. Each component is described below.
\subsection{Graph Generator}\label{subsec:graph_gen}
Given the output of the query mechanism for a query, the Graph Generator creates a directed graph with $5$ types of nodes: authoring entities, referenced entities, issues, events and documents. Directed edges are used by Composer to update source node representations using destination nodes. We design the topology with the main goal of capturing the representations of events, issues and referenced entities that reflect author's opinion about them. We add edges from issues/events to author's documents but omit the other direction as our main goal is to contextualize issues/events using author's opinions.\\
\indent Bidirectional edges from authors to their Wikipedia articles, tweets, press releases and perspectives, from issues to their background description, events and from events to news articles describing them are added. Uni-directional edges from events to tweets and press releases, from issues to author perspectives and from referenced entities to the documents that mention them are added. An example graph is shown in Fig. \ref{fig:test_graph}.
\begin{figure}[!bth]
    \centering
    \includegraphics[width=4.7cm]{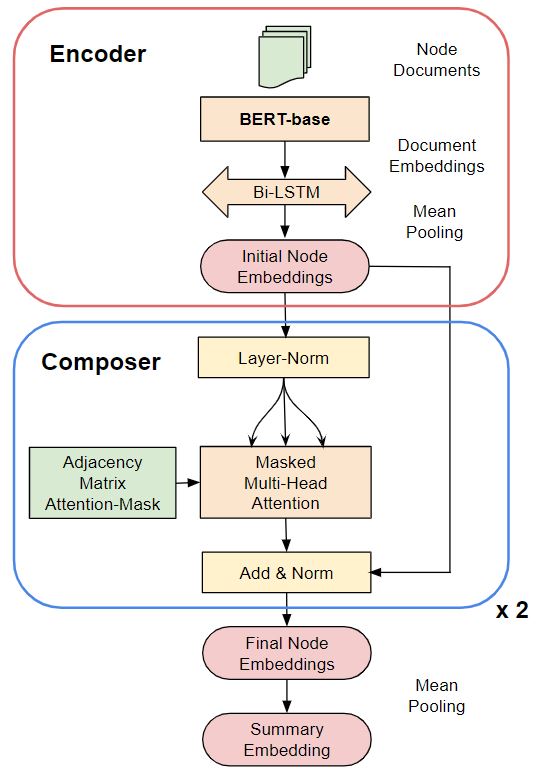}
    \caption{Encoder-Composer Architecture}
    \label{fig:comp_reader}
\end{figure}
\subsection{Encoder}
Encoder is used to compute the initial node embeddings. It consists of BERT followed by a Bi-LSTM. For each node, it takes a sequence of documents as input. The documents are ordered temporally. The output of Encoder is a single embedding of dimension $d_m$ for each node. Given a node $\mathcal{N}$ = \{$D_1$, $D_2$, \ldots{}, $D_d$\} consisting of $d$ documents, for each document $D_i$, contextualized embeddings of all the tokens are computed using BERT. Token embeddings are computed sentence-wise to avoid truncating long documents. Then, token embeddings of each document are mean-pooled to get the document embeddings $\mathcal{\vec{N}}^{bert}$ = \{$\vec{D_1}^{bert}$, $\vec{D_2}^{bert}$, \ldots{}, $\vec{D_d}^{bert}$\} where $\vec{D_i}^{bert} \in \mathbb{R}^{1 \times d_m}$, $d_m$ is the dimension of a BERT token embedding. The sequence $\vec{\mathcal{N}}^{bert}$ is passed through a Bi-LSTM to obtain an output sequence $\vec{E}$ = \{$\vec{e_1}$, $\vec{e_2}$, \ldots{}, $\vec{e_d}$\}, $\vec{e_i} \in \mathbb{R}^{1 \times h}$, where $h/2$ is the hidden dimension of the Bi-LSTM, we set $h = d_m$ in our model. Finally, the output of Encoder is computed by mean-pooling the sequence $\vec{E}$. We use BERT-base-uncased model in our experiments where $d_m=h=768$. Initial node embeddings of all the document nodes are set to Encoder output of the documents themselves. For authoring entity nodes, their Wikipedia descriptions, tweets, press releases and perspective documents are passed through Encoder. For issue nodes, background description of the issue is used. For event nodes, all the news articles related to the event are used. For referenced entities, all documents that mention the entity are used.
\subsection{Composer}
Composer is a transformer-based graph attention network (GAT) followed by a pooling layer. We use the transformer encoding layer proposed by \citet{transformer-2017}, without the position-wise feed forward layer, as graph attention layer. Position-wise feed forward layer is removed as in contrast with sequence-to-sequence prediction tasks, nodes in a graph usually have no ordering relationship between them. Adjacency matrix of the graph is used as the attention mask. Self-loops are added for all nodes so that updated representation of the node also depends on its previous representation. Composer module uses $l=2$ graph attention layers in our experiments. Composer module generates updated node embeddings $\mathbb{U} \in \mathbb{R}^{n \times d_m}$ and a summary embedding $\mathbb{S} \in \mathbb{R}^{1 \times d_m}$ as outputs. The output dimension of node embeddings is $768$. Equations that describe Composer unit are:

\small
\begin{equation}\label{eqn:gat}
\centering
\begin{multlined}
    \mathbb{E} \in \mathbb{R}^{d_m \times n}, \mathcal{A} \in \{0, 1\}^{n \times n}\hfill\\
    \mathbb{G} = LN(\mathbb{E})\hfill\\
    Q = W_q^T \mathbb{G}, K = W_k^T \mathbb{G}, V = W_v^T \mathbb{G}\hfill\\
    M = \frac{Q^{T} K}{\sqrt{d_k}}, M = mask(M, \mathcal{A})\hfill\\
    \mathbb{O} =  M V^{T}, \mathbb{U} = W_o^T \mathbb{O} + \mathbb{E}\hfill\\
    \mathbb{S} = \textit{mean-pool}(\mathbb{U})\hfill
\end{multlined}
\end{equation}
\normalsize
\noindent where $n$ is number of nodes in the graph, $d_m$ is the dimension of a BERT token embedding, $d_k$, $d_v$ are projection dimensions, $n_h$ is number of attention heads used and $Q \in \mathbb{R}^{n_h \times d_k \times n}$, $K \in \mathbb{R}^{n_h \times d_k \times n}$, $V \in \mathbb{R}^{n_h \times d_v \times n}$, $\mathbb{O} \in \mathbb{R}^{n_h d_v \times n}$, $ M \in \mathbb{R}^{n_h \times n \times n}$. $W_q \in \mathbb{R}^{d_m \times n_h d_k}$, $W_k \in \mathbb{R}^{d_m \times n_h d_k}$, $W_v \in \mathbb{R}^{d_m \times n_h d_v}$ and $W_o \in \mathbb{R}^{n_h d_v \times d_m}$ are weight parameters to be learnt. $\mathbb{E} \in \mathbb{R}^{d_m \times n}$ is the output of the encoder. $\mathcal{A} \in \{0, 1\}^{n \times n}$ is the adjacency matrix. We set $n_h=12$ and $d_k=d_v=64$.

\section{Learning Tasks}\label{sec:task}
We design two learning tasks to train the Compositional Reader model: \textit{Authorship Prediction} and \textit{Referenced Entity Prediction}. Both the tasks are intuitively designed to train the model to learn the association between the author node representation and the language used by the particular author. These tasks are two variations of link prediction over the graph. The tasks are detailed below.
\subsection{Authorship Prediction}\label{subsec:auth_pred}
Authorship Prediction is designed as a binary classification task. In this task, the model is given a graph generated by the graph generator in subsection \ref{subsec:graph_gen}, an author node and a document node. The task is to predict whether or not the document was authored by the input author.\\
\indent Intuition behind this learning task is to enable our model to learn differentiating between: 1) language of an author's first-person discourse vs. third person discourse in news articles, 2) language of an author vs. language used by other authors and 3) language of an author in context of one issue vs. in context of other issues. The model sees documents by the author in the graph and learns to decide whether or not the input document is by the same author and talking about the same issue.

\begin{table}[!bth]
    \centering
    \resizebox{180pt}{!}{
        \begin{tabular}{lcccc}
        \hline
            \textbf{Model} & \textbf{IS Acc} & \textbf{IS F1} & \textbf{OS Acc} & \textbf{OS F1}\\\hline
            \multicolumn{3}{l}{\textbf{Authorship Prediction}} & & \\\hline
            BERT Adap. &  $93.01$ & $92.31$ & $95.56$ & $95.20$ \\
            Comp. Reader &  $99.49$ & $99.47$ & $99.42$ & $99.39$ \\\hline
            
            \multicolumn{3}{l}{\textbf{Reference Entity Prediction}} &  & \\\hline
            BERT Adap. &  $76.57$ & $75.21$ & $76.26$ & $73.67$\\
            Comp. Reader &  $78.52$ & $77.51$ & $78.98$ & $78.62$ \\\hline
        \end{tabular}
    }
    \caption{Learning Tasks In-Sample \& Out-Sample Results on Test Data. Acc.denotes Accuracy. F1 Score for the Positive Class is Reported.}
    \label{tab:learn_task_res}
\end{table}

\noindent \textbf{Data} Training data for the task was created as follows: for a particular author-issue pair, we obtain a data graph similar to Fig. \ref{fig:test_graph} using the query mechanism in subsection \ref{subsec:query_mech}. To create a positive data sample, we sample a document $d_i$ authored by the entity $a_i$ and remove the edges between the nodes $a_i$ and $d_i$. Negative samples were designed carefully in $3$ batches to align with our above task objectives. In the first batch, we sample news article nodes from the same graph. In the second batch, we obtain tweets, press releases and perspectives of the same author but from a different issue. In the third batch, we sample documents related to the same issue but from other authors. We generate $421,284$ samples in total, with $252,575$ positive samples and $168,709$ negative samples. We randomly split the data into training set of $272,159$ samples, validation set of $73,410$ samples and test set of $75,715$ samples.\\
\noindent \textbf{Architecture} We concatenate the initial and final node embeddings of the author, document and also the summary embedding of the graph to obtain inputs to the fine-tuning layers for Authorship Prediction task. We add one hidden layer of dimension $384$ before the classification layer.\\
\noindent \textbf{Out-sample Evaluation} We perform out-sample experiments to evaluate generalization capability to unseen author data. We train the model on training data from two-thirds of politicians and test on the test sets of others. Results are shown in Tab. \ref{tab:learn_task_res}.\\
\noindent \textbf{Graph Trimming} We perform graph trimming to make the computation tractable on a single GPU. We randomly drop $80\%$ of the news articles, tweets and press releases that are not related to the event to which $d_i$ belongs. We use graphs with $200$-$500$ nodes and batch size of $1$.
\subsection{Referenced Entity Prediction}\label{subsec:ref_ent_pred}
This is also a binary classification task. Given a data graph, a document node with a masked entity and a referenced entity node the graph, the task is to predict whether the referenced entity is same as the masked entity. Intuition behind this learning task is to enable our model to learn the correlation between the language of the author in the document and the masked entity. For example, in context of recent Donald Trump's impeachment hearing, consider the sentence `X needs to face the consequences of their actions'. Depending upon the author, X could either be `\textit{Donald Trump}' or `\textit{Democrats}'. Learning to understand such correlations by looking at other documents from the same author is effective in capturing meaningful author representations.\\
\noindent \textbf{Data} To create training data, we sample a document from the data graph. We mask the most frequent entity in the document with a generic $<$\texttt{ENT}$>$ token. We remove the link between the masked entity and the document in the data graph. We sample another referenced entity from the graph to generate a negative example. We generated $252,578$ samples for this task, half of them positive. They were split into $180,578$ training samples, validation and test sets of $36,400$ samples each.\\
\noindent \textbf{Architecture} We use fine-tuning architecture similar to Authorship Prediction on top of Compositional Reader for this task as well. We keep separate fine-tuning parameters for each task as they are fundamentally different prediction problems. Compositional Reader is shared. We apply graph trimming for this task as well. We also perform out-sample evaluation for this learning task.\\
\noindent \textbf{Results} Performance of the BERT Adaptation baseline and Compositional Reader model are shown in Tab \ref{tab:learn_task_res}. On Authorship Prediction, out-sample performance doesn't drop for either model. This shows the usefulness of our graph formulation which allows the models to learn linguistic nuances. On Referenced Entity Prediction, F1 score for our model improves from $77.51$ from in-sample to $78.62$ on out-sample while BERT adaptation baseline's F1 drops slightly from $75.21$ to $73.67$

\section{Evaluation}\label{sec:eval}
\begin{table*}[!tbph]
    \centering
    \resizebox{375pt}{!}{
        \begin{tabular}{lcccccc}
            \hline
            \textbf{Model} & \thead{\bf Paraphrase\\\bf All Grades} & \thead{\bf Paraphrase\\\bf A/F Grades} & \thead{\bf NRA\\\bf Val Acc} & \thead{\bf NRA\\\bf Test Acc}  & \thead{\bf LCV\\\bf Val Acc} & \thead{\bf LCV\\\bf Test Acc} \\\hline
             BERT & $41.55\%$ & $38.52\%$ & $55.93 \pm 0.72$ & $54.83 \pm 1.79$ & $54.28 \pm 0.31$ & $52.63 \pm 1.21$ \\
             
             BERT Adap. & $37.54\%$ & $42.62\%$ & $71.23 \pm 3.93$ & $69.95 \pm 3.33$ & $60.58 \pm 1.56$ & $59.09 \pm 1.77$ \\
             
             Encoder &  $56.16\%$ & $48.36\%$ & $83.95 \pm 1.24$ & $81.34 \pm 0.86$ & $65.10 \pm 0.46$ & $63.42 \pm 0.35$ \\
             
             Comp. Reader &  $63.32\%$ & $63.93\%$ & $84.19 \pm 0.98$ & $81.62 \pm 1.23$ & $65.55 \pm 1.33$ & $62.24 \pm 0.56$ \\\hline
        \end{tabular}
    }
    \caption{ Results of \textit{Grade Paraphrase} and \textit{Prediction} tasks. Acc denotes Accuracy, NRA and LCV denote Grade Prediction tasks. Mean $\pm$ Std. Dev for $5$ random seeds for Grade Prediction showing statistical significance. }
    \label{tab:quant_eval}
\end{table*}
\begin{table*}[!tbh]
\centering
    \resizebox{375pt}{!}{
        \begin{tabular}{cccccccccc}
        \hline
        \textbf{Session} & \textbf{Majority Class} (\%) & \multicolumn{2}{c}{\textbf{Accuracy (\%)}} & \multicolumn{2}{c}{\textbf{Precision (\%)}} & \multicolumn{2}{c}{\textbf{Recall (\%)}} & \multicolumn{2}{c}{\textbf{F1 (\%)}} \\ 
                &                     & \textbf{NW-GL}            & \textbf{CR}              & \textbf{NW-GL}            & \textbf{CR}               & \textbf{NW-GL}          & \textbf{CR}             & \textbf{NW-GL}         & \textbf{CR}           \\ \hline
        \textbf{106}     & 83.23               & 85.04            & 85.65           & 91.89            & 91.67            & 90.22           & 91.27          & 91.05         & 91.47        \\ 
        \textbf{107}     & 85.78               & 87.62            & 88.30           & 90.12            & 89.48            & 95.37           & 97.17          & 92.67         & 93.16        \\ 
        \textbf{108}     & 87.02               & 92.03            & 92.27           & 93.46            & 93.52            & 97.59           & 97.83          & 95.48         & 95.32        \\ 
        \textbf{109}     & 83.57               & 85.42            & 87.23           & 88.38            & 88.39            & 93.84           & 97.33          & 91.49         & 92.65        \\ \hline
        \textbf{Average} & 84.90               & 87.53            & 88.36           & 90.96            & 90.77            & 94.26           & 95.90          & 92.67         & 93.15        \\ \hline
        \end{tabular}
    }
    \caption{Roll Call Prediction Results. NW-GL represents the best performing model of \citet{roll-call-2019} as replicated by us using their official implementation. CR represents Compositional Reader results. The improvements are statistically significant as per McNemar's test.}
    \label{tab:roll_call_res}
\end{table*}

We evaluate our model and pre-training tasks in a systematic manner using several quantitative tasks and qualitative analysis. Quantitative evaluation includes \textit{Grade Paraphrase} task, \textit{Grade Prediction} on \textit{National Rifle Association (NRA)} and \textit{League of Conservation Voters (LCV)} grades data followed by \textit{Roll Call Vote Prediction} task. Qualitative evaluation includes entity-stance visualization for issues and Opinion Descriptor Generation. We compare our model's performance to BERT representations, the BERT adaptation baseline and representations from the Encoder module. Baselines and the evaluation tasks are detailed below. Further evaluation tasks are in the appendix.

\subsection{Baselines}\label{subsec:lp_baseline}
\textbf{BERT:} We compute the results obtained by using pooled BERT representations of relevant documents for each of the quantitative tasks. Details of the chosen documents and the pooling procedure is described in the relevant task subsections. We chose BERT-base over BERT-large due to the complexity of running the learning tasks on embedding dimension $768$ vs $1024$. A bigger embedding dimension results in lesser context (lesser number of nodes in the graph). 

\noindent \textbf{Encoder Representations:} We compare the performance of our model to the results obtained by using initial node embeddings generated from the Encoder for each of the quantitative tasks.

\noindent \textbf{BERT Adaptation Model:} We design a BERT adaptation baseline for the learning tasks. BERT adaptation architecture is same as the Encoder of the Compositional Reader model. While Encoder's parameters are trained via back-propagation through the Composer, BERT adaptation model is directly trained on learning tasks. In BERT adaptation, once we generate the data graph, we pass the mean-pooled sentence-wise BERT embeddings of the node documents through a Bi-LSTM. We mean-pool the output of Bi-LSTM to get node embeddings. We use fine-tuning layers on top of thus obtained node embeddings for both the learning tasks. BERT Adaptation baseline allows us to showcase the importance of our proposed training tasks via comparison with BERT-base representations. It also demonstrates the usefulness of Composer.

\subsection{Grade Paraphrase Task}
\textit{National Rifle Association} (NRA) assigns letter grades (A+, A, \ldots{}, F) to politicians based on candidate questionnaire and their gun-related voting. We evaluate our representations on their ability to predict these grades. We collected the historical data of politicians' NRA grades from \href{https://everytown.org/nra-grades-archive/}{everytown.org}.

In \textit{Grade Paraphrase} task, we evaluate our representations directly \textit{without} training on the NRA data. Grades are divided into two classes: grades including and above \textit{B+} are in positive class and grades from \textit{C+} to \textit{F} are clustered into negative. We formulate representative sentences for them:
\begin{itemize}[noitemsep]
    \item POSITIVE: \textit{I strongly support the NRA}
    \item NEGATIVE: \textit{I vehemently oppose the NRA}
\end{itemize}
For each politician, we obtain data graph for the issue \textit{guns}. We input the data graph to Compositional Reader model and use the node embeddings of the author politician ($\vec{n}_{auth}$), issue \textit{guns} ($\vec{n}_{guns}$) and referenced entity \textit{NRA} ($\vec{n}_{NRA}$). For some politicians, $\vec{n}_{NRA}$ is not available as they have not referenced \textit{NRA} in their discourse. We just use $\vec{n}_{auth}$ and $\vec{n}_{guns}$ for them. We compute BERT-base embeddings for the representative sentences to obtain $\vec{pos}_{NRA}$ and $\vec{neg}_{NRA}$. We mean-pool the three embeddings $\vec{n}_{auth}$, $\vec{n}_{guns}$ and $\vec{n}_{NRA}$ to obtain $\vec{n}_{stance}$. We compute cosine similarity of $\vec{n}_{stance}$ with $\vec{pos}_{NRA}$ \& $\vec{neg}_{NRA}$. Politician is assigned the higher similarity class.

\indent We compare our model's results to BERT-base, BERT adaptation and Encoder embeddings. For BERT-base, we compute $\vec{n}_{stance}$ by mean-pooling the sentence-wise BERT embeddings of tweets, press releases and perspectives of the author on all events related to the issue \textit{guns}. Results are shown in Tab. \ref{tab:quant_eval}. Compositional Reader achieves $63.32\%$ accuracy. Encoder embeddings get $56.16\%$. Mean-pooled BERT-base embeddings get $41.55\%$. Using node embeddings from BERT adaptation model yields $37.54\%$. When we evaluate using only `A'/`F' grades, we obtain $63.93\%$ accuracy for Compositional Reader, $48.36\%$ for Encoder, $42.62\%$ for BERT adaptation and $38.52\%$ for BERT-base.

\subsection{Grade Prediction Task}
\noindent \textbf{NRA Grades} This is designed as a $5$-class classification task for grades \{\textit{A, B, C, D \& F}\}. We train a simple feed-forward network with one hidden layer. The network is given $2$ inputs $\vec{n}_{auth}$ \& $\vec{n}_{guns}$. When $\vec{n}_{NRA}$ is available for an author, we set $\vec{n}_{guns}$ = \textit{mean}($\vec{n}_{NRA}$, $\vec{n}_{guns}$). The output is a binary prediction.

\indent We perform $k=10$-fold cross-validation for this task. We repeat the entire process for $5$ random seeds and report the results with confidence intervals. We perform this evaluation for BERT-base, BERT adaptation, Encoder and Compositional Reader. To compute $\vec{n}_{auth}$ for BERT-base, we mean-pool the sentence-wise embeddings of all author documents on \textit{guns}. For $\vec{n}_{guns}$, we use the background description document of issue \textit{guns}. Results on the test set are in Tab. \ref{tab:quant_eval}.

\noindent \textbf{LCV Grades} This is similar to NRA Grade Prediction task. This is a $4$-way classification task. \textit{League of Conservation Voters} (LCV) assigns a score ranging between $0$-$100$ to each politician depending upon their environmental voting activity. We segregate politicians into $4$  classes ($0-25$, $25-50$, $50-75$, $75-100$). We obtain input to the prediction model by concatenating $\vec{n}_{auth}$ and $\vec{n}_{environment}$. We use same fine-tuning architecture as NRA Grade Prediction task.

\indent Results of Grade Prediction task are shown in Tab. \ref{tab:quant_eval}. On \textit{NRA Grade Prediction}, which is a $5$-way classification task, our model achieves an accuracy of $81.62 \pm 1.23$ on the test set. Our model outperforms BERT representations by $26.79 \pm 3.02$ absolute points on the test set. On \textit{LCV Grade Prediction} task which is a $4$-way classification, our model achieves $9.61 \pm 1.77$ point improvement over BERT representations.

\subsection{Roll Call Vote Prediction Task}
This task was proposed in \citet{roll-call-2019}. We skip the finer details of the task for brevity. The task aims to predict the voting behaviour of US politicians on roll call votes. Given the bill texts and voting history of the politicians, the aim is to predict future voting patterns of the politicians. We inject our politician author embeddings from Compositional Reader model to improve the performance on the task. We input all the politician first-person discourse from our data to compute politician author embeddings using Compositional Reader model. We use these embeddings to initialize the legislator embeddings in their news-augmented glove model, which is their best performing model. We use the data splits provided in their official implementation. We use their code to reproduce their results. Results are shown in Tab. \ref{tab:roll_call_res}.

\subsection{Qualitative Evaluation}
\begin{figure}[!ht]
  \centering
  \subfigure[Individual Stances]{\includegraphics[scale=0.29]{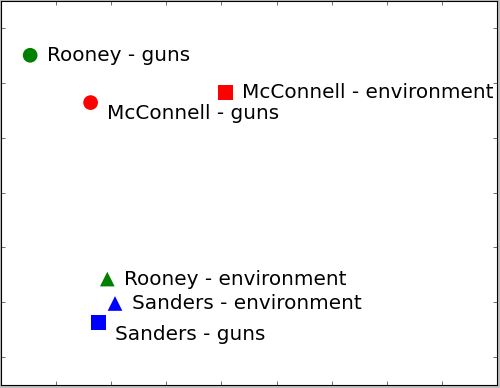}}\hfill
  \subfigure[Issue \textit{Guns}]{\includegraphics[scale=0.29]{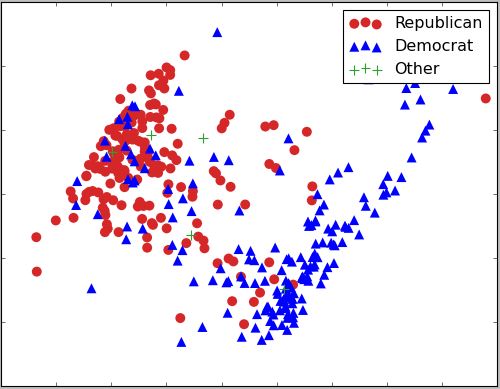}}\hfill
  \caption{PCA Visualizations of Politician Embeddings}
  \label{fig:pca_viz}
\end{figure}

\noindent \textbf{Politician Visualization} We perform Principle Component Analysis (PCA) on issue embeddings ($\vec{n}_{issue}$) of politicians obtained using the same method as in NRA Grade prediction. We show one such interesting visualization in Fig. \ref{fig:pca_viz}. Sen. McConnell, a Republican who expressed right-wing views on both \textit{environment} and \textit{guns}. Sen. Sanders, a Democrat that expressed left-wing views on both. Rep. Rooney, a Republican who expressed right-wing views on \textit{guns} but left-wing views on \textit{environment}. Fig. \ref{fig:pca_viz} demonstrates that this information is captured by our representations. Additional such visualizations are included in the appendix.

\begin{table*}[!tbp]
    \centering
    \resizebox{460pt}{!}{%
        \begin{tabular}{l l l l}
        \hline
            \textbf{Issue} & \textbf{Opinion Descriptors} & \textbf{Issue} & \textbf{Opinion Descriptors}\\\hline
             Mitch McConnell & Republican & Nancy Pelosi  & Democrat \\\hline
             \textit{abortion} & fundamental, hard, eligible, embryonic, unborn & \textit{abortion} & future, recent, scientific, technological, low\\
             \textit{environment} & achievable, more, unobjectionable, favorable, federal & \textit{environment} & forest, critical, endangered, large, clear\\
             \textit{guns} & substantive, meaningful, outdone, foreign, several & \textit{guns} & constitutional, ironclad, deductible, unlawful, fair\\
             \textit{immigration} & federal, sanctuary, imminent, address, comprehensive & \textit{immigration} & immigrant, skilled, modest, overall, enhanced\\\hline
             Donald Trump & Republican & Joe Biden  & Democrat \\\hline
             \textit{guns} & terrorist, public, ineffective, huge, inevitable, dangerous & \textit{guns} & banning, prohibiting, ban, maintaining, sold\\
             \textit{immigration} & early, dumb, birthright, legal, difficult & \textit{taxes} & progressive, economic, across-the-board, annual, top\\ \hline
        \end{tabular}
    }
    \caption{Opinion Descriptor Labels for Politicians. They show the most representative adjectives used by the politicians in context of each issue. }
    \label{tab:opinion_descs}
\end{table*}

\begin{table}[!tbp]
    \centering
    \resizebox{!}{32pt}{
        \begin{tabular}{lc|lc}
            \hline
            \textbf{Model} & \textbf{Accuracy} & \textbf{Model} & \textbf{Accuracy} \\\hline
            Comp.Reader & $63.32\%$ & & \\
            w/o Tweets & $63.32\%$ & Only Tweets & $40.11\%$\\
            w/o Press & $63.04\%$ & Only Press  & $55.87\%$\\
            w/o Persp. & $59.31\%$ & Only Persp. & $60.74\%$\\\hline
        \end{tabular}
    }
    \caption{Ablation Study on \textit{Grade Paraphrase} task for various types of documents}
    \label{tab:ablation_study}
\end{table}

\noindent \textbf{Issue Visualization} We present visualization of politicians on the issue \textit{guns} in Fig. \ref{fig:pca_viz}. We observe that \textit{guns} tends to be a polarizing issue. This shows that our representations are able to effectively capture relative stances of politicians. We observe that issues that have traditionally had clear conservative vs liberal boundaries such as \textit{guns} \& \textit{abortion} are more polarized compared to issues that evolve with time such as \textit{middle-east} \& \textit{economic-policy}. Visualization for issue \textit{abortion} is in the appendix.

\subsection{Opinion Descriptor Generation}
\indent This task demonstrates a simple way to interpret our contextualized representations as natural language descriptors. It is an unsupervised qualitative evaluation task. We generate opinion descriptors for authoring entities for specific issues. We use the final node embedding of the issue node ($\vec{n}_{issue}$) for each politician to generate opinion descriptors.

Inspired from \citet{larn-2019}, we define our candidate space for descriptors as the set of adjectives used by the entity in their tweets, press releases and perspectives related to an issue. Although \citet{larn-2019} uses verbs as relationship descriptor candidates, we opine that adjectives describe opinions better. We compute the representative embedding for each descriptor by mean-pooling the contextualized embeddings of that descriptor from all its occurrences in the politician's discourse. This is the one of the key differences with prior descriptor generation works such as \citet{larn-2019} and \citet{rmn-2016}. They work in a static word embedding space. But, our embeddings are contextualized and also reside in a higher dimensional space. In an unsupervised setting, this makes it more challenging to translate from distributional space to natural language tokens. Hence, we restrict the candidate descriptor space more than \citet{larn-2019} and \citet{rmn-2016}. We rank all the candidate descriptors according to cosine similarity of its representative embedding with the vector $\vec{n}_{issue}$. 

We present some of the results in Tab. \ref{tab:opinion_descs}. In contrast to \citet{rmn-2016} and \citet{larn-2019}, our model doesn't need the presence of both the entities in text to generate opinion descriptors. This is often the case in first person discourse. Results are shown in table \ref{tab:opinion_descs}.

\subsection{Ablation Study}

\indent Further, we investigate the importance of various components. We perform ablation study over various types of documents on the \textit{NRA Grades Paraphrase} task. the results are shown in Tab. \ref{tab:ablation_study}. They indicate that \textit{perspectives} are most useful while \textit{tweets} are the least useful documents for the task. As \textit{perspectives} are summarized ideological leanings of politicians, it is intuitive that they are more effective for this task. Tweets are informal discourse and tend to be very specific to a current event, hence they are not as useful for this task.  

\section{Conclusion}\label{sec:conc}
We tackle the problem of \textit{understanding politics}, i.e., creating unified representations of political figures capturing their views and legislative preferences, directly from raw political discourse data originating from multiple sources. We propose the Compositional Reader model that composes multiple documents in one shot to form a unified political entity representation,  while capturing the real-world context needed for representing the interactions between these documents. 

We  evaluate our model on several qualitative and quantitative tasks. We outperform BERT-base model on both types of tasks. Our qualitative evaluation demonstrate that our representations effectively capture nuanced political information.

\section*{Acknowledgements}\label{sec:ackn}
We thank Shamik Roy, Nikhil Mehta and the anonymous reviewers for their insightful comments.  This work was partially supported by an NSF
CAREER award IIS-2048001.

\bibliography{refs}
\bibliographystyle{acl_natbib}

\begin{appendices}


\section{Event Examples}\label{sec:events_app}
\indent In this section, we provide examples of events that were identified by our event identification heuristic. For each automatically extracted event, we observe that the news headlines with in the cluster usually describe the same real world event. The span of each event is $10$ days at most. Hence, the assumption that the events with in each issue are non-overlapping is a reasonable relaxation of reality. We made event segregated document data available for future research along with our code. Examples are shown in Tab \ref{tab:event_eg}.

\begin{table*}[!tbh]
    \centering
    \resizebox{450pt}{!}{
    \begin{tabular}{l|l}
    \hline
    \multicolumn{2}{c}{\textbf{Issue - Economic Policy}} \\
    \hline
        \textbf{Event \#1: Donald Trump's Tax Proposal Release} & \textbf{Event \#2: Obama's Economy Speech} \\ \hline
            Donald Trump to Propose Tax Breaks on ‘Pocketbook’ Issues in Economic Plan & Obama Economy Speech: Why The President's Plan Won't Get Past Republicans \\
            Trump's economic plan aims to please both corporations and working families & U.S. Is 'Through The Worst Of Yesterday's Winds,' Obama Says \\
            Donald Trump Looks to Steady His Campaign With New Economic Speech & Obama Blames Five Years of a Bad Economy on "Phony Scandals" and "Distractions"\\
            Trump to outline economic plan in Detroit & Obama tries to offset current scandals by recycling talking points on economy \\
            Clinton to dismiss Trump's economic plan as a 'friends and family discount' & Obama at Knox College: ‘Washington has taken its eye off the ball\\
            OPINION: Trump agenda looks like more of the same & Obama Says Private Capital Should Take Lead Mortgage Role\\
            Clinton to dismiss Trump's economic plan as a 'friends and family discount' & Why Obama might tap Summers for Fed despite harsh criticism from left\\
            Trump tries to right his campaign, talking of tax cuts & Obama: Growing income inequality ‘defining challenge’ of this generation \\\hline
        \end{tabular}
    }
    \caption{Examples of extracted events}
    \label{tab:event_eg}
\end{table*}

\section{Reproducibility}\label{sec:repr_app}
We use seeds (set to $4056$ for both tasks) for both random example generation and training neural networks. For fine-tuning layers of learning tasks we initialize the models using Xavier uniform \citep{xavier-init-2010} initialization with gain=$1.0$. We optimize the parameters using Stochastic Gradient Descent with an initial learning rate=$0.0075$ and momentum=$0.4$. We used $4$ Nvidia GeForce GTX $1080$ Ti GPUs with $12$ GB memory and linux servers with $64$ GB RAM for our experiments. CPU RAM and GPU memory are the main bottlenecks for training the model. It takes $80$ hours to train authorship prediction for $5$ epochs and $14$ hours to train referenced entity prediction task for the same. Generating test results for both tasks together takes $3$ hours. We use a batch size of $1$ for both training and evaluation. For NRA Grade Prediction task we use $5$ random seeds: \{$5, 7, 11, 13, 17$\} and report mean and standard deviation. The encoder-composer architecture is made up of $8.26$M parameters, encoder consisting of $3.54M$ and composer $4.72M$. Due to long training time, the only hyper-parameter we experimented with is the graph size. We retained as many nodes as possible without exceeding GPU memory ($500$ nodes).\\
\indent We divide the $3,640$ queries into $151$ batches of $24$ queries each ($3$ politicians $\times$ $8$ issues) and $1$ batch of $16$ queries ($2$ politicians $\times$ $8$ issues). Train, val and test data examples are generated for each query batch.  For Authorship Prediction and Referenced Entity Prediction tasks, Compositional Reader model is trained on one batch for $5$ epochs, the best parameters are chosen according to the validation performance of that batch and we proceed to training on future batches. Politicians are ordered randomly when generating queries.

\subsection{Data Collection}
We collected data from $5$ sources: Wikipedia, Twitter, \href{https://www.ontheissues.org}{ontheissues.org}, \href{https://www.allsides.com}{allsides.com} and ProPublica Congress API. We scraped articles from Wikipedia related to all the politicians in focus. We collected tweets from \href{https://github.com/alexlitel/congresstweets}{Congress Tweets} and \citet{trump-tweets-2019}. We used a set of hand build gold hashtags to separate them by issues. They are shown at the end of this document. We collected all news articles related to the $8$ issues in focus from \href{https://www.allsides.com}{allsides.com}. We collected press releases from Propublica API using key word search. We use issue names as keywords. We only maintain pointers to processed tweet and text data in data releases. All social media text analyzed is by public political figures, not private citizens.

\begin{table*}[!tbh]
    \centering
    \resizebox{450pt}{!}{
        \begin{tabular}{l}
            \hline
            \textbf{Guns:} \\
            \#endgunviolence, \#guncontrol, \#gunviolence, \#nra, \#gunsafety, \#assaultweaponsban, \#gunsense, \#marchforourlives, \#parkland, \#hr3435,\\ \#nationalwalkoutday, \#disarmhate, \#guncontrolnow, \#backgroundchecks, \#nationalschoolwalkout, \#lasvegas, \#elpaso, \#keepamericanssafe,\\ \#gunrights, \#erpoact, \#lasvegasshooting, \#gunreform, \#hr1112, \#parklandstrong, \#elpasostrong, \#massshootings, \#parklandstudentsspeak, \#hr8 \\\hline
            
            \textbf{Taxes:}\\ \#GOPTaxScam, \#TaxReform, \#TaxAndJobsAct, \#taxreform, \#goptaxscam, \#taxcutsandjobsact, \#taxday, \#taxcuts,\\ \#smallbusinessweek, \#economy, \#maga, \#billionairesfirst, \#gopbudget, \#goptaxplan, \#goptaxbill, \#tax, \#taxscam, \#trumptax\\\hline
            
            \textbf{Immigration:}\\ \#FamiliesBelongTogether, \#Immigration, \#MuslimBan, \#daca, \#familiesbelongtogether, \#dreamers, \#immigration, \#protectdreamers,\\ \#dreamactnow, \#muslimban, \#heretostay, \#keepfamiliestogether, \#protectthedream, \#defenddaca, \#immigrants, \#familyseparation,\\ \#nomuslimbanever, \#immigrant, \#nobannowall, \#borderwall, \#refugeeswelcome, \#endfamilydetention, \#protectfamilies, \#DACA, \#refugees\\\hline
            
            \textbf{Abortion:}\\ \#ProChoice, \#ProLife, \#Abortion, \#prolife, \#abortion, \#marchforlife, \#prochoice, \#theyfeelpain, \#bornaliveact, \#paincapable, \#hr36,\\ \#roevwade, \#unplanned, \#defundpp, \#life, \#standwithnurses, \#endinfanticide, \#righttolife, \#infanticide, \#ppsellsbabyparts\\\hline
            
            \textbf{LGBTQ Rights:}\\ \#LGBTQ, \#LGBT, \#Homophobia, \#lgbtq, \#lgbt, \#equalityact, \#pridemonth, \#hr5, \#nationalcomingoutday, \#lgbtqequalityday, \#loveislove,\\ \#lgbthistorymonth, \#transgender, \#letkidslearn, \#trans, \#comingoutday, \#marriageequality, \#protecttranstroops, \#lgbtqhistorymonth,\\ \#defundconversiontherapy, \#transban, \#otd, \#prideinprogress, \#nycpride, \#protecttranskids, \#transrightsarehumanrights,\\ \#transdayofremembrance, \#loveisthelaw, \#rfra, \#bathroombill\\\hline
            
            \textbf{Middle-East:}\\ \#MiddleEast, \#Iran, \#Israel, \#iran, \#israel, \#syria, \#middleeast, \#iraq, \#russia, \#northkorea, \#irandeal, \#jordan, \#hezbollah, \#gaza,\\ \#isis, \#hamas, \#terror, \#jihad, \#violence, \#barbarism, \#palestinians, \#jewish, \#antisemitism, \#saudiarabia, \#iranian, \#lebanon, \#turkey,\\ \#jerusalem, \#iranprotests, \#israeli, \#freeiran, \#sanctions, \#supportisrael, \#egypt, \#terrorism\\\hline
            
            \textbf{Environment:}\\ \#ActOnClimate, \#ClimateChange, \#GreenNewDeal, \#climatechange, \#actonclimate, \#greennewdeal, \#climateactionnow, \#parisagreement,\\ \#climatecrisis, \#earthday, \#climatefriday, \#climate, \#climatestrike, \#climateaction, \#cleanenergy, \#climatechangeisreal, \#environment,\\ \#oceanclimateaction, \#cleanair, \#climatechangeimpactsme, \#cleanwater, \#globalwarming, \#renewableenergy, \#worldenvironmentday,\\ \#climateemergency, \#peopleoverpolluters, \#greenjobs, \#climatejustice, \#solar, \#environmentaljustice, \#cleanpowerplan, \#todaysclimatefact,\\ \#sealevelrise, \#bigoil, \#climatecrisiscountdown, \#stopextinction, \#cleanercars, \#climatecosts, \#cutmethane, \#chamberofcarbon,\\ \#climatesolutions,  \#amazonrainforest, \#hurricanemaria, \#climatesecurityisnationalsecurity, \#protectcleanwater, \#renewables, \#offfossilfuels,\\ \#columbiaenergyexchange, \#climatesolutionscaucus\\\hline
            
        \end{tabular}
    }
    \caption{Gold Hashtag Set used to collect Politicians' Tweets for Issues}
    \label{tab:my_label}
\end{table*}

\section{Evaluation Tasks}\label{sec:eval_app}
\subsection{Grade Prediction Additional Ablation}
\begin{figure}
    \begin{subfigure}{}
        \centering
        \includegraphics[width=6.5cm]{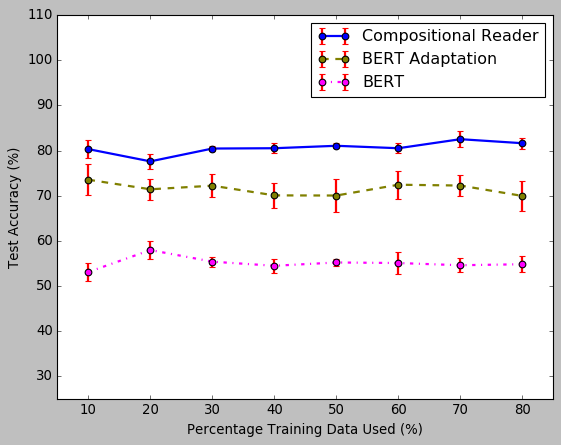}
        \caption{NRA Grade Prediction: Data \% vs Test Acc}
        \label{fig:nra_test_acc}
    \end{subfigure}
    
    \begin{subfigure}{}
        \centering
        \includegraphics[width=6.5cm]{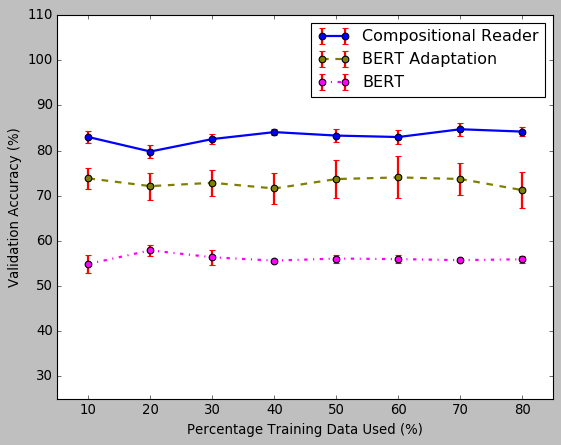}
        \caption{NRA Grade Prediction: Data \% vs Val Acc}
        \label{fig:nra_grade_pred_res_val}
    \end{subfigure}
    
    \begin{subfigure}{}
        \centering
        \includegraphics[width=6.5cm]{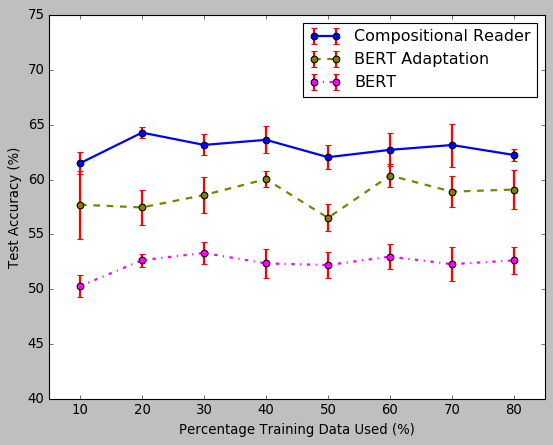}
        \caption{LCV Grade Prediction: Data \% vs Test Acc}
        \label{fig:lcv_test_res}
    \end{subfigure}
    
    \begin{subfigure}{}
        \centering
        \includegraphics[width=6.5cm]{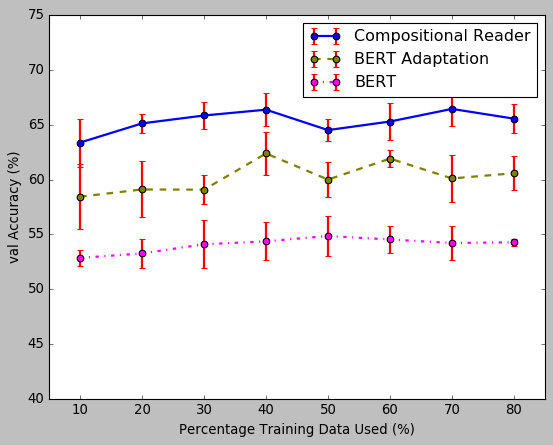}
        \caption{LCV Grade Prediction: Data \% vs Val Acc}
        \label{fig:lcv_val_res}
    \end{subfigure}
\end{figure}

For Grade Prediction task, we perform experiments by training the model on a fraction of the data. We monitor the validation and test performances with change in training data percentage. We observe that, in general, the gap between Compositional Reader model and the BERT baseline widens with increase in training data. It hints that our representation likely captures more relevant information for this task. Results are shown in figure \ref{fig:nra_test_acc}.

\subsection{Opinion Descriptor vs. RMN and LARN}
\citet{larn-2019} and \citet{rmn-2016} both take a set of documents and entity pairs as inputs and generate relationship descriptors for the entity pairs in an unsupervised setting. They are both trained in an encoder-decoder style training process in an unsupervised manner. Given new text with an entity pair, they generate $d$ descriptor embeddings that are used to rank candidate descriptors. \citet{rmn-2016} uses entire vocabulary space while \citet{larn-2019} uses $500$ most frequent verbs.

\indent In contrast, our model doesn't need the presence of both the entities in text to generate opinion descriptors. This often tends to be the case in tweets and press releases as they are generated directly by the author (first-person discourse). Our model is also capable of summarizing over multiple documents and generating descriptors for several referenced entities and issues at once while they deal with one entity-pair at a time.

\clearpage
\section{Additional Visualizations}\label{sec:addn_viz_app}
\begin{figure}[!ht]
    \centering
    \begin{subfigure}{}
        \centering
        \includegraphics[width=6.25cm]{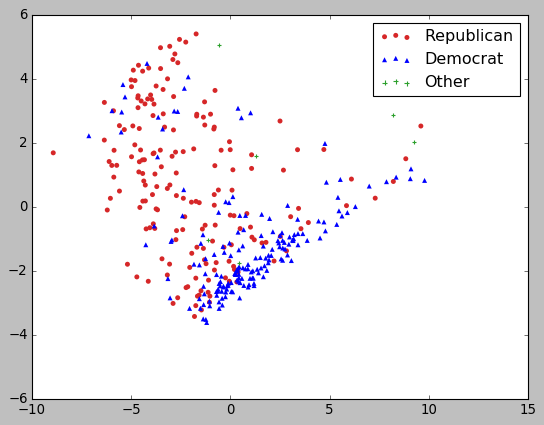}
        \caption{Reps vs Dems on issue \textit{immigration}}
        \label{fig:pca_immigration}
    \end{subfigure}
    
    \begin{subfigure}{}
        \centering
        \includegraphics[width=6.25cm]{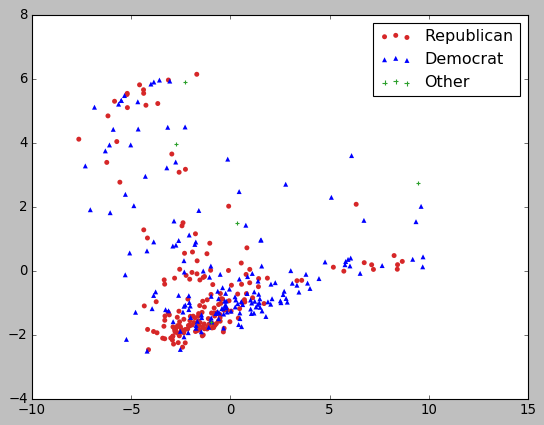}
        \caption{Reps vs Dems on issue \textit{taxes}}
        \label{fig:pca_taxes}
    \end{subfigure}
    
    \begin{subfigure}{}
        \centering
        \includegraphics[width=6.25cm]{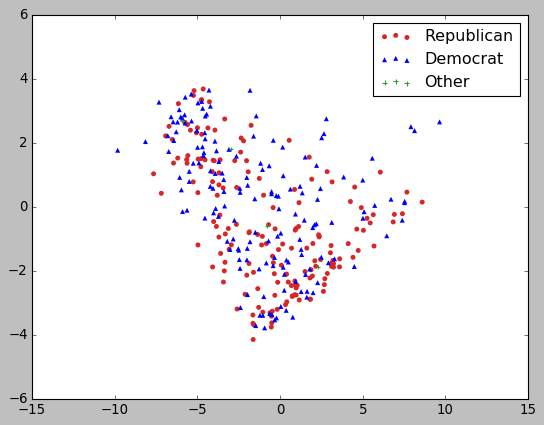}
        \caption{Reps vs Dems on issue \textit{middle-east}}
        \label{fig:pca_middle-east}
    \end{subfigure}
    
    \begin{subfigure}{}
        \centering
        \includegraphics[width=6.25cm]{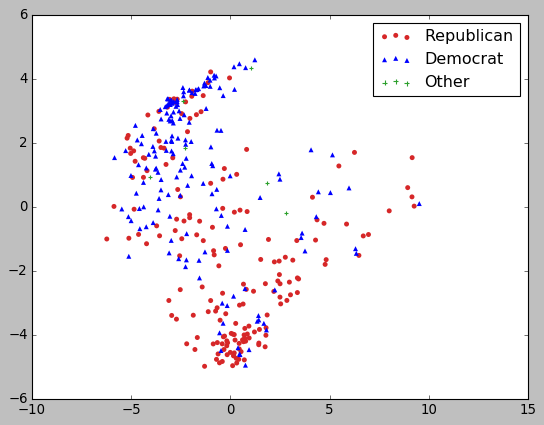}
        \caption{Reps vs Dems on issue \textit{abortion}}
        \label{fig:pca_abortion}
    \end{subfigure}
\end{figure}

\begin{figure}
    \begin{subfigure}{}
        \centering
        \includegraphics[width=6.5cm]{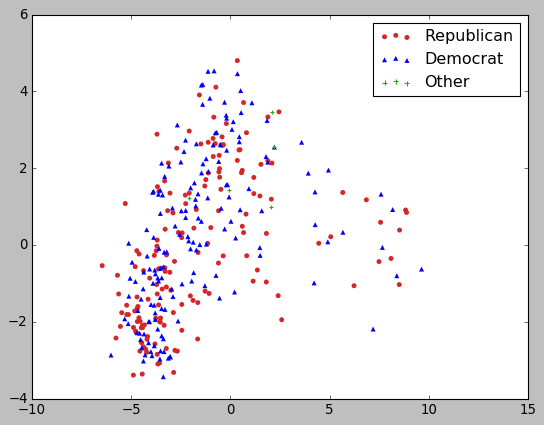}
        \caption{Reps vs Dems on issue \textit{economic-policy}}
        \label{fig:pca_economic-policy}
    \end{subfigure}
    
    \begin{subfigure}{}
        \centering
        \includegraphics[width=6.5cm]{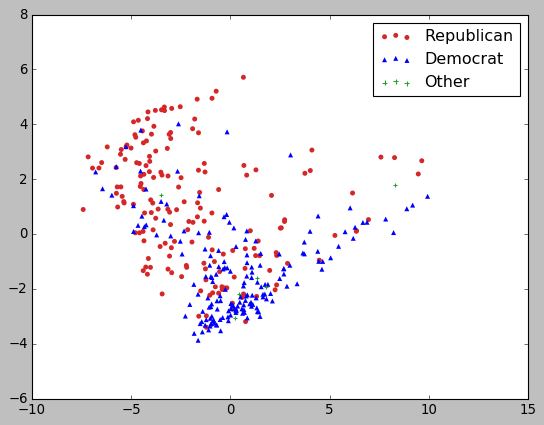}
        \caption{Reps vs Dems on issue \textit{environment}}
        \label{fig:pca_environment}
    \end{subfigure}
    
    \begin{subfigure}{}
        \centering
        \includegraphics[width=6.5cm]{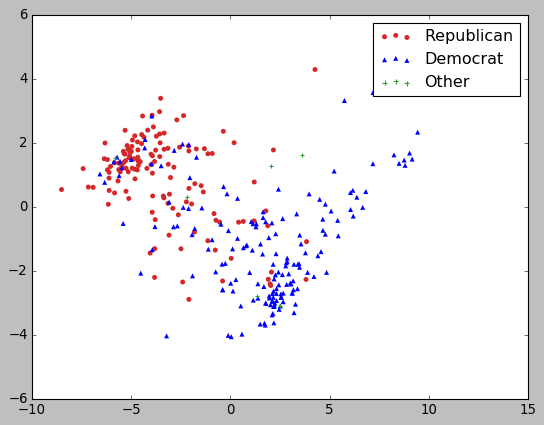}
        \caption{Reps vs Dems on issue \textit{LGBTQ Rights}}
        \label{fig:pca_lgbtq-rights}
    \end{subfigure}
    
    \begin{subfigure}{}
        \includegraphics[width=6.5cm]{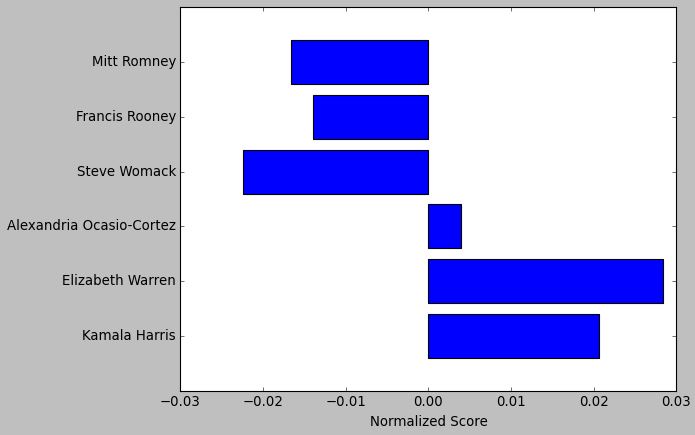}
        \caption{Normalized Agreement Score of Politicians with statement: \textit{Believes in common-sense approach to reforming gun control.}}
        \label{fig:para_bar_plot}
    \end{subfigure}
\end{figure}

\begin{figure}
    \centering
    \begin{subfigure}{}
        \includegraphics[width=6.5cm]{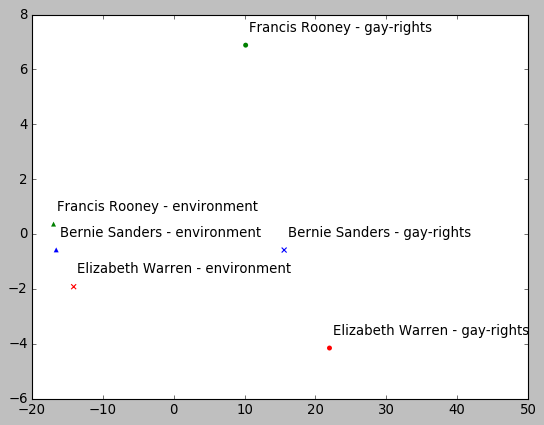}
        \caption{Comparison of Politician Stances on Issues}
        \label{fig:rooney_warren_sanders}   
    \end{subfigure}
    
    \begin{subfigure}{}
        \includegraphics[width=6.5cm]{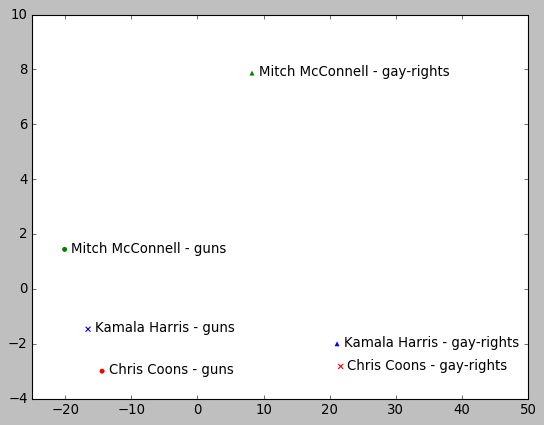}
        \caption{Comparison of Politician Stances on Issues}
        \label{fig:coons_kamals_mcconnell}
    \end{subfigure}

    \centering
    \begin{subfigure}{}
        \includegraphics[width=6.5cm]{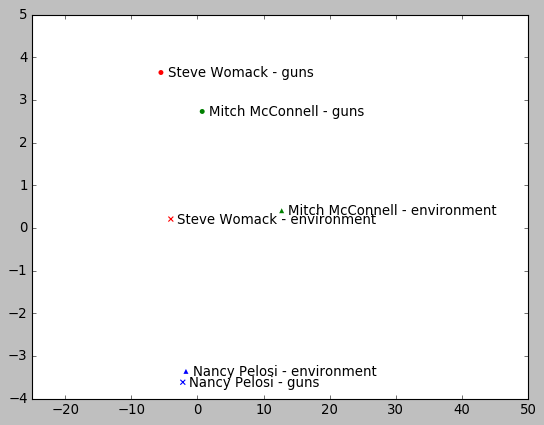}
        \caption{Comparison of Politician Stances on Issues}
        \label{fig:womack_mcconnell_pelosi}
    \end{subfigure}
    
\end{figure}

\end{appendices}

\end{document}